\documentclass[conference]{IEEEtran}
\IEEEoverridecommandlockouts
\usepackage{cite}
\usepackage{amsmath,amssymb,amsfonts}
\usepackage{algorithmic}
\usepackage{graphicx}
\usepackage{textcomp}
\usepackage{xcolor}
\usepackage{multirow}
\usepackage[subrefformat=parens]{subcaption}
\usepackage{comment}
\usepackage{url}
\usepackage{threeparttable}

\DeclareMathOperator*{\argmax}{arg\,max}
\def\BibTeX{{\rm B\kern-.05em{\sc i\kern-.025em b}\kern-.08em
    T\kern-.1667em\lower.7ex\hbox{E}\kern-.125emX}}
\usepackage{caption}
\usepackage{fancyhdr}
\renewcommand\IEEEkeywordsname{Keywords}
\begin{document}

\title{Feature Selective Likelihood Ratio Estimator for Low- and Zero-frequency N-grams}

\author{\IEEEauthorblockN{\small Masato Kikuchi}
\IEEEauthorblockA{\small \textit{Dept. Comput. Sci.,} \\
\textit{Grad. Sch. Eng.} \\
\textit{Nagoya Institute of Technology} \\
Aichi, 466-8555, Japan \\
kikuchi@nitech.ac.jp}
\and
\IEEEauthorblockN{\small Mitsuo Yoshida}
\IEEEauthorblockA{\small \textit{Dept. Dept. Comput. Sci. Eng.} \\
\textit{Toyohashi University of Technology} \\
Aichi, 441-8580, Japan \\
yoshida@cs.tut.ac.jp} \\
\and
\IEEEauthorblockN{\small Kyoji Umemura}
\IEEEauthorblockA{\small \textit{Dept. Comput. Sci. Eng.} \\
\textit{Toyohashi University of Technology} \\
Aichi, 441-8580, Japan \\
umemura@tut.jp} \\
\and
\IEEEauthorblockN{\small Tadachika Ozono}
\IEEEauthorblockA{\small \textit{Dept. Comput. Sci.,} \\
\textit{Grad. Sch. Eng.} \\
\textit{Nagoya Institute of Technology} \\
Aichi, 466-8555, Japan \\
ozono@nitech.ac.jp}
}

\maketitle

\begin{abstract}
In natural language processing (NLP), the likelihood ratios (LRs) of N-grams are often estimated from the frequency information.
However, a corpus contains only a fraction of the possible N-grams, and most of them occur infrequently.
Hence, we desire an LR estimator for low- and zero-frequency N-grams.
One way to achieve this is to decompose the N-grams into discrete values, such as letters and words, and take the product of the LRs for the values.
However, because this method deals with a large number of discrete values, the running time and memory usage for estimation are problematic.
Moreover, use of unnecessary discrete values causes deterioration of the estimation accuracy.
Therefore, this paper proposes combining the aforementioned method with the feature selection method used in document classification, and shows that our estimator provides effective and efficient estimation results for low- and zero-frequency N-grams.
\end{abstract}


\section{Introduction}
Likelihood ratios (LRs) are defined as the ratio of probability distributions and are used in various applications.
We estimate them for actual use, and the estimation performance can be an important determinant of the effectiveness of the applications.
In natural language processing (NLP), LRs for strings or word sequences are often estimated using the frequency information obtained from a corpus.
One way to estimate an LR is to estimate each probability distribution as the relative frequency $\widehat{p}_*(x)$ and take their ratio:
\begin{align}
r_{\text{MLE}}(x) = \frac{\widehat{p}_{\text{nu}}(x)}{ \widehat{p}_{\text{de}}(x)}, \quad
\widehat{p}_*(x) = \frac{f_*(x)}{n_*}, \quad * \in \{{\text{de, nu}}\}.
\end{align}
$x=\langle t_1, t_2, \ldots, t_{\text{N}} \rangle$ represents a sequence of N letters or words, which is called an N-gram.
$t_k,\ 1 \le k \le {\text{N}}$ is the $k$-th letter or word that comprises $x$.
$f_*(x)$ is the observed frequency of $x$ obtained from the probability distribution with density $p_*(x)$ and $n_*=\sum_x f_*(x)$.
Although the above method is simple, the estimate may be unreasonably large for low frequencies, and a small difference in frequencies significantly varies the estimate.
There are many types of language elements such as N-grams, and most of them occur infrequently.
Therefore, when estimating LRs from the frequencies, we often face these problems.

The estimator was proposed to address low-frequency problems~\cite{Kikuchi:19}, which is defined as
\begin{align}
\widehat{r} (x) &= \left(\frac{f_{\text{de}}(x)}{n_{\text{de}}} + \lambda \right)^{-1} \times \frac{f_{\text{nu}}(x)}{n_{\text{nu}}},
\end{align}
where $\lambda$ is a parameter described later.
Unlike the ``indirect'' estimator, which involves probability distribution estimation, this estimator directly estimates the LR by solving a squared loss minimization problem.
The regularization parameter $\lambda\ (\ge 0)$ introduced in the minimization problem plays an important role in mitigating low-frequency problems.
To confirm it, we show estimation examples in TABLE~\ref{tab:esti_example}.
First, there is a considerable difference between $f_*(x_{\text{a}})$ and $f_*(x_{\text{b}})$, whereas $r_{\text{MLE}}(x_{\text{a}})$ and $r_{\text{MLE}}(x_{\text{b}})$ are both 50, which is a large value.
Here, $f_{\text{nu}}(x_{\text{b}})=1$ may be a coincidence occurrence.
Therefore, a frequency-based ``reliability'' should be reflected in the estimates.
On the other hand, $\widehat{r}(x_{\text{a}})$ is 47.6, which is close to 50, whereas $\widehat{r}(x_{\text{b}})$ is 8.3, which is far lower than 50.
Therefore, $\widehat{r}(x)$ reflects the reliability.
Next, even though the difference between $f_{\text{nu}}(x_{\text{b}})$ and $f_{\text{nu}}(x_{\text{c}})$ is only one, $r_{\text{MLE}}(x _{\text{c}})$ varies greatly from 50 to 100.
However, because $f_*(x_{\text{b}})$ and $f_*(x_{\text{c}})$ are both low frequencies, the estimates should be robust in this situation.
The difference between $\widehat{r}(x_{\text{b}})$ and $\widehat{r}(x_{\text{c}})$ is small because $\widehat{r}(x)$ underestimates depending on the low frequency.
Thus, $\widehat{r}(x)$ is a robust estimator for low-frequencies.
Finally, we focus on $x_{\text{d}}$, $r_{\text{MLE}}(x_{\text{d}})=\widehat{r}(x_{\text{d}})=0$ because $f_{\text{nu}}(x_{\text{d}})$ is zero.
This means that even $\widehat{r}(x)$ cannot calculate informative estimates for zero-frequency N-grams, which are not observed in a corpus.
\begin{table}
\centering
\begin{threeparttable}[tb]
\caption{Examples of LR Estimation\tnote{$\dagger$}.}
\label{tab:esti_example}
	\begin{tabular}{ c  r  r  r  r  r  r } \hline
		 \multicolumn{1}{ c }{N-gram} & \multicolumn{4}{ c }{Observed freq.} & \multicolumn{1}{ c }{\multirow{2}{*}{$r_{\text{MLE}}(x)$}} & \multicolumn{1}{ c }{\multirow{2}{*}{$\widehat{r}(x)$}} \\ \cline{2-5}
		$x$ & \multicolumn{1}{ c }{$n_{\text{de}}$} & \multicolumn{1}{ c }{$f_{\text{de}}(x)$} & \multicolumn{1}{ c }{$n_{\text{nu}}$} & \multicolumn{1}{ c }{$f_{\text{nu}}(x)$} & &  \\ \hline
		$x_{\text{a}}$ & $10^{7}$ & 2,000 & $10^{4}$ & 100 & 50 & 47.6 \\
		$x_{\text{b}}$ & $10^{7}$ & 20 & $10^{4}$ & 1 & 50 & 8.3 \\
		$x_{\text{c}}$ & $10^{7}$ & 20 & $10^{4}$ & 2 & 100 & 16.7 \\
		$x_{\text{d}}$ & $10^{7}$ & 6 & $10^{4}$ & 0 & 0 & 0 \\ \hline
	\end{tabular}
	\begin{tablenotes}
\item[$\dagger$] $\lambda$ of $\widehat{r}(x)$ is $10^{-5}$.
\end{tablenotes}
\end{threeparttable}
\end{table}

As mentioned earlier, there are many types of language elements in a corpus, and most of them occur infrequently.
Moreover, there are many zero-frequency elements.
In NLP applications (e.g., machine translation systems and information retrieval systems), zero-frequency strings or search queries, which are not contained in the training data, are often given as input, and informative estimates are required even in such cases.
To address these cases, we desire an LR estimator to provide informative estimates for zero-frequency N-grams.

A simple way to deal with zero frequencies is to decompose $x$ into discrete values $\{t_k\}_{k=1}^{\rm N}$ and approximate $r(x)$ by the product of $r(t_k)$s.
This general treatment for $t_k$s also applies to naive Bayes classifiers.
Additionally, by using the estimator~\cite{Kikuchi:19}, we can obtain the robust estimator $\widehat{r}(t_k)$ for low-frequency N-grams.
However, the treatment of $t_k$s shows following problems.
First, to treat $t_k$s individually, we must implicitly assume statistical independence between $t_k$s.
This assumption often does not hold in practice, and may deteriorate the accuracy of LR estimation.
Second, decomposing $x$ into $t_k$s forces us to deal with a large variety of $t_k$s. 
There are many $t_k$s which are unnecessary or harmful for estimation.
Therefore, treating all $t_k$s blindly reduces the estimation accuracy and efficiency.
To mitigate these problems, we combined the feature selection method for document classification using the method described above.
In other words, we propose to select only those estimates from the available $t_k$s which are useful for efficient estimation.
In our experiments, we predict the occurrence contexts of named entities from a corpus using LRs.
We compare our estimator with estimators that use all $t_k$s and show that our estimator achieves efficient LR estimation while maintaining the same or better prediction accuracy.

\section{Related Work}

Many methods have been proposed for probability estimation in NLP to deal with low and zero frequencies ~\cite{Chen:99}.
They are called smoothing techniques, which discount a certain amount from the probability estimates of observed events and distribute them to the estimates of unobserved events.
In this framework, the probabilities for low-frequency events are underestimated, and those for zero-frequency events are estimated to be slightly greater than zero.
Although smoothing techniques are not directly applicable to LR estimation, it is possible to estimate the probability distributions using these techniques and determine their ratio.
However, the estimates obtained by this approach were shown to be impractical~\cite{Kikuchi:19}, and this fact motivated our research.

Indirect estimators that involve probability distribution estimation were shown to yield large estimation errors~\cite{Hardle:12}.
Therefore, direct LR estimation methods without going through distribution estimation were proposed~\cite{Sugiyama:08,Kanamori:09}.
These direct estimation methods are for LRs defined on continuous spaces.
In contrast, we deal with N-grams obtained from discrete sample spaces, and LRs estimated from their frequencies are also defined in discrete spaces.
Kikuchi et al.~\cite{Kikuchi:19} made uLSIF~\cite{Kanamori:09}, which is a direct estimation method, applicable to estimation of LRs defined in discrete spaces.
In this estimator, the regularization parameter introduced in an optimization scheme can provide robust estimates from low frequencies.
However, even the estimator ~\cite{Kikuchi:19} cannot calculate informative estimates from zero frequencies.

Because a document contains a huge variety of words, word handling is often problematic in document classification.
Most of the words occur infrequently and do not contribute to classification.
On the contrary, some words induce misclassification.
For these reasons, feature selection methods have been proposed to eliminate noisy, uninformative, and redundant words.
Feature selection methods can be categorized into four types depending on how to generate a feature (word) subset: filter, wrapper, embedding, and hybrid models.
Among these, the filter models are often used for their efficiency and effectiveness.
The filter models~\cite{Mladenic:99,Yang:97,Galavotti:00} select a word subset $\Theta \subset V$ for training using score functions.
$V$ denotes the vocabulary of the training set.
Our estimator has a regularization parameter to be tuned; therefore, we combine the estimator with a filter model that is computationally efficient.

\section{Preliminaries}

We explain the direct LR estimation method~\cite{Kikuchi:19} and the feature selection method~\cite{Mladenic:99,Yang:97,Galavotti:00}, which are necessary to introduce our proposed estimator in Section~\ref{sec:our_method}.

\subsection{Direct Likelihood Ratio Estimation Method}\label{sec:uLSIF}

Let $D \subset \mathbb{U}$ be a data domain, where $\mathbb{U}$ is a set of $v$ discrete elements, also called a finite alphabet in information theory.
Suppose we have two i.i.d. samples
\begin{align}
\{x_{i}^{\text{de}}\}_{i=1}^{n_{\text{de}}} \overset{\text{i.i.d.}}{\sim} p_{\text{de}}(x), \quad \{x_{j}^{\text{nu}}\}_{j=1}^{n_{\text{nu}}} \overset{\text{i.i.d.}}{\sim} p_{\text{nu}}(x),
\end{align}
where the element $x$ is a discrete value such as a word (sequence) or a letter (sequence), and $v$ is the number of element types that can exist.
Following previous studies, we assume that the probability density $p_{\text{de}}(x)$ satisfies the following condition:
\begin{align}
p_{\text{de}}(x) > 0 \quad {\text{for\ all\ }} x \in D,
\end{align}
which allows us to define LRs for all $x$.
Here, we estimate the following LR directly from the samples $\{x_{i}^{\text{de}}\}_{i=1}^{n_{\text{de}}}$ and $\{x_{i}^{\text{nu}}\}_{i=1}^{n_{\text{nu}}}$ without estimating the probability distributions.
\begin{align}
r(x) = \frac{p_{\text{nu}}(x)}{p_{\text{de}}(x)}.
\end{align}

Unconstrained least-squares importance fitting (uLSIF)~\cite{Kanamori:09} is a direct LR estimation method using a squared loss minimization process.
uLSIF models $r(x)$ as the linear sum
\begin{align}
\widehat{r}(x) = \sum_{l=1}^{b} \beta_l \varphi_l (x),
\end{align}
where $\mbox{\boldmath $\beta$}=(\beta_1, \beta_2, \ldots, \beta_{b})^{\mathrm{T}}$ are the parameters to be learned from the samples, and $\{\varphi_l\}_{l=1}^{b}$ are the basis functions that take non-negative values.
Note that $b$ and $\{\varphi_l\}_{l=1}^{b}$ are independent of the samples $\{x_{i}^{\text{de}}\}_{i=1}^{n_{\text{de}}}$ and $\{x_{i}^{\text{nu}}\}_{i=1}^{n_{\text{nu }}}$, respectively.
The original uLSIF exploits the structure of continuous sample spaces by using basis functions based on Gaussian kernels.
However, the sample spaces we deal with are discrete, and thus, Gaussian kernels are not effective.
Therefore, we substitute the basis functions $\{\delta_l\}_{l=1}^{v}$ proposed by Kikuchi et al.~\cite{Kikuchi:19}.
\begin{align}
\label{eq:phi}
\delta_l(x)= 
\begin{cases}
    1 & (x = x_{(l)}) \\
    0 & (x \ne x_{(l)})
  \end{cases}
\end{align}
where index $l$ specifies a particular element from among $v$ element types that can exist.
That is, $x_{(l)}$ denotes the $l$-th element out of the $v$ element types.
Although $\{\delta_l\}_{l=1}^{v}$ do not capture the relationship between elements, introducing them into uLISF has the advantage that the optimal solution can be obtained analytically.
Substituting Eq.(\ref{eq:phi}) into $\widehat{r}(x_{(m)}),\ 1 \le m \le v$, we obtain
\begin{align}
\label{eq:linear_model}
\widehat{r}(x_{(m)}) = \sum_{l=1}^{v} \beta_l \delta_l (x_{(m)}) = \beta_m.
\end{align}
In uLISF, the parameters $\mbox{\boldmath $\beta$}$ are learned to minimize the squared loss of $\widehat{r}(x_{(m)})$ and the true LR $r(x_{(m)})$.
The optimization problem is defined as\footnote{For the derivation of Eq.(\ref{eq:uLSIF}), see the original paper of uLSIF~\cite{Kanamori:09}.}
\begin{align}
\label{eq:uLSIF}
\min_{\mbox{\boldmath $\beta$} \in \mathbb{R}^v} \left[\frac{1}{2} \mbox{\boldmath $\beta$}^{\mathrm{T}} \widehat{\mbox{\boldmath $H$}} \mbox{\boldmath $\beta$} - \widehat{\mbox{\boldmath $h$}}^{\mathrm{T}} \mbox{\boldmath $\beta$} + \frac{\lambda}{2} \mbox{\boldmath $\beta$}^{\mathrm{T}} \mbox{\boldmath $\beta$}\right],
\end{align}
where $\mathbb{R}^v$ is a real $v$-dimensional space.
The penalty term $\frac{\lambda}{2} \mbox{\boldmath $\beta$}^{\mathrm{T}} \mbox{\boldmath $\beta$}$ is introduced to regularize $\mbox{\boldmath $\beta$}$.
$\lambda\ (\ge 0)$ is a regularization parameter, and $\mbox{\boldmath $\beta$}^{\mathrm{T}}\mbox{\boldmath $\beta$}/2$ is an $\ell_2$regularization term.
$\widehat{\mbox{\boldmath $H$}}$ is a $v \times v$ matrix, and its $(l, l')$-th element is defined as
\begin{align}
\label{eq:H_hat}
\widehat{H}_{l,l'} = \frac{1}{n_{\text{de}}}\sum_{i=1}^{n_{\text{de}}} \delta_l(x_i^{\text{de}}) \delta_{l'}(x_i^{\text{de}})
= \begin{cases}
	\frac{f_{\text{de}}(x_{(l)})}{n_{\text{de}}} & (l=l') \\
	0 & (l \neq l')
	\end{cases}
\end{align}
where $f_{*}(x_{(l)}),\ * \in \{\text{de, nu}\}$ is the frequency of $x_{(l)}$ obtained from the probability distribution with density $p_{*}(x)$.
From the above, $\widehat{\mbox{\boldmath $H$}}$ becomes a diagonal matrix.
$\widehat{\mbox{\boldmath $h$}}$ is a $v$-dimensional vector, and its $l$-th element is defined as
\begin{align}
\label{eq:h_hat}
\widehat{h}_l = \frac{1}{n_{\text{nu}}} \sum_{j=1}^{n_{\text{nu}}} \delta_l (x_j^{\text{nu}}) = \frac{f_{\text{nu}} (x_{(l)})}{n_{\text{nu}}}.
\end{align}
Eq.(\ref{eq:uLSIF}) is an unconstrained quadratic programming problem whose solution can be obtained analytically as follows:
\begin{align*}
\widetilde{\mbox{\boldmath $\beta$}} (\lambda)=(\widehat{\mbox{\boldmath $H$}} + \lambda \mbox{\boldmath $1$}_{v})^{-1} \widehat{\mbox{\boldmath $h$}},
\end{align*}
where $\mbox{\boldmath $1$}_{v}$ is a $v$-dimensional vector with all ones.
From Eqs.(\ref{eq:linear_model}), (\ref{eq:H_hat}), and (\ref{eq:h_hat}), the solution to Eq.(\ref{eq:uLSIF}), is as follows:
\begin{align}
\label{eq:solution}
\widehat{r}(x_{(m)}) = \left(\frac{f_{\text{de}} (x_{(m)})}{n_{\text{de}}} + \lambda \right)^{-1} \times \frac{f_{\text{nu}} (x_{(m)})}{n_{\text{nu}}}.
\end{align}
In the original uLSIF, the solution could be negative, and the values were rounded to zero, taking into account the non-negativity of LRs.
However, because the above equation is always non-negative, $\widetilde{\beta}_m(\lambda)$ is the final solution.

In Eq.(\ref{eq:solution}), the regularization parameter $\lambda\ (\ge 0)$ makes the estimate smaller and more robust.
This equation is derived from the squared loss minimization with $\ell_2$-regularization and corrects the denominator with a constant corresponding to the regularization strength.
When $\lambda=0$, this is equivalent to $\widehat{p}_{\rm nu}(x) / \widehat{p}_{\rm de}(x)$, where $\widehat{p}_*(x)$ is the relative frequency.

\subsection{Feature Selection Method for Document Classification}

Naive Bayes classifiers, which are the leading machine learning algorithms, are probabilistic classifiers based on the strong independence assumption and Bayes' theorem.
Suppose we are given a document $d$ represented by a vector of $\text{N}$ words $\langle t_1, t_2, \ldots, t_{\text{N}}\rangle$.
Here, $t_k,\ 1 \le k \le {\text{N}}$ indicates a word in the $k$-th position from the beginning of the document.
When $C$ is a class variable and $c$ is a value taken by $C$, the classifiers that solve the problem of classifying $d$ into the appropriate class can be formulated as
\begin{align}
\widehat{c}(d) &= \argmax_{c \in C} p(c) \prod_{k=1}^{\text{N}} p(t_k \mid c),
\end{align}
where $\widehat{c}(d)$ is the class label under which $d$ is classified.
The classifiers treat each word $t_k$ in $d$ individually and approximate the conditional probability $p(d \mid c)$ by the product of $p(t_k \mid c)$s.
This approximation is based on the assumption that the occurrence of $t_k$ is conditionally independent of other words under $c$.
However, this often does not hold in practice and is known to deteriorate the classification accuracy.
Furthermore, handling a huge variety of words reduces the computational efficiency.

Feature selection methods were proposed to mitigate the above-mentioned problems.
These methods select a word subset $\Theta$ that is useful for classification from the vocabulary $V$ of the training data.
By using $\Theta$ for training, classifiers improved the classification accuracy and efficiency.
Feature selection methods define score functions for words and use scores as a criterion for selecting words that $\Theta$ should contain.
In this study, we used the three score functions, the expected cross entropy for text (CET) ~\cite{Mladenic:99}, chi-square statistic ~\cite{Yang:97}, and {\textit{GSS coefficient}}~\cite{Galavotti:00}, which have been commonly used.
The details of the functions are described in Section~\ref{sec:comparison_methods}.

\section{The Proposed Likelihood Ratio Estimator}\label{sec:our_method}

Suppose we estimate the following LR for the feature vector $x=\langle t_1, t_2, \ldots, t_{\text{N}} \rangle$, where $t_k,\ 1\le k\le {\text{N}}$ is a discrete value such as a letter or word, and $x$ is a sequence of $\text{N}$ discrete values, which is called an N-gram.
\begin{align}
r(x) = \frac{p_{\text{nu}}(x)}{p_{\text{de}}(x)}.
\end{align}
A simple way to estimate $r(x)$ approximates each probability distribution using the relative frequency and taking their ratio.
However, because N-grams are language elements, most of them are low- or zero-frequency elements.
In particular, as N increases, the relative frequencies in the denominator and numerator become zero, making it difficult to calculate informative estimates.

A simple solution for this problem is to treat each $t_k$ in $x$ individually and approximate $r(x)$ by the product of $r(t_k)$s.
This approximation is based on the assumption that the occurrence of $t_k$ is statistically independent of the occurrence of other discrete values.
Additionally, by using the estimator in Section \ref{sec:uLSIF} to estimate $r(t_k)$, we can obtain robust estimates even for low-frequency N-grams.
However, this assumption often does not hold in practice, which may deteriorate the accuracy of the LR estimation.
Furthermore, handling $t_k$ increases the running time and memory usage for the estimation.

Therefore, we introduce a feature selection method to select only the values with discriminative power from the training data.
Our proposed estimator is formulated as
\begin{align}
\label{eq:proposed}
r_{\text{ours}}(x) &= \prod_{k=1}^{\text{N}} \widetilde{r}(t_k)^{w_{k(m)}}, \\
\widetilde{r}(t_k) &= \left\{\frac{f_{\text{de}}(t_k) + 1}{n_{\text{de}} + 2} + \lambda \right\}^{-1} \frac{f_{\text{nu}}(t_k) + 1}{n_{\text{nu}} + 2},
\end{align}
where $\lambda\ (\ge 0)$ is a regularization parameter.
Note that if we use the original frequency of $t_k$, the estimate of $x$ that contains at least one $t_k$ with $f_{\text{nu}}(t_k)=0$ would be zero.
To avoid this problem, we use the corrected frequency by adding 1 and 2 to $f_*(t_k)$ and $n_*$, respectively.
Our estimator realizes feature selection by the following weight $w_{k(m)}$, which considers both the value type $m$ and occurrence position $k$:
\begin{align}
\label{eq:weight}
w_{k(m)}= 
\begin{cases}
    1 & (t_{k(m)} \in \Theta_k) \\
    0 & (t_{k(m)} \not \in \Theta_k)
  \end{cases}
\end{align}
where $t_{k(m)}$ is the discrete value of the $m$-th type at the $k$-th position of an N-gram, and $\Theta_k$ is a subset of $V_k$, which is a set of discrete values that can exist at position $k$.
The subset size $|\Theta_k|$ is a hyperparameter.
In Eq.(\ref{eq:proposed}), the proposed estimator uses selected values for estimation by assigning a weight of one, and ignores other by assigning a weight of zero.
The score functions in Section \ref{sec:comparison_methods} are used to determine $w_{k(m)}$.

\section{Experiments}

We predict word 10-grams\footnote{We regarded a string separated by a space as a word.
In constructing 10-grams, we did not perform any special pre-processing such as lemmatization or stop words removal.
} on the left of named entities (location names and persons' names, tagged as LOC and PER, respectively) using LRs.
We have three reasons for conducting the experiments.
First, N-grams have a wide variety of types; however, many of them occur infrequently.
This property facilitates validation for effectiveness of feature selection.
Second, the difficulty of feature selection differs between entity types.
In the left N-grams of PER, honorific titles and nouns that mean position or occupation tend to occur on the left of entities.
Therefore, selecting them significantly affects the prediction accuracy.
In contrast, in the left N-grams of LOC, prepositions tend to occur on the left of entities; however, they also tend to occur in other contexts.
Hence, words at other positions should be considered for prediction.
Two types of left N-grams help clarify the behavior of our estimator.
Third, the left N-grams are uniquely determined, which allows for a quantitative evaluation.
We verified the effectiveness of our estimator in terms of prediction accuracy, running time, and memory usage.
Additionally, we prepared several score functions and clarified the differences in the behavior of each function.

\subsection{Experimental Environment}

The experimental environment is shown below.
\begin{itemize}
\item OS: Windows 10 Pro
\item Processor: Intel Xeon W3520 @ 2.67GH
\item Memory: 16.0 GB
\item Perl: v5.30.2
\end{itemize}

\subsection{Datasets and Conditions}

We created datasets using the 1987 edition of the Wall Street Journal corpus.
First, we randomly distributed the articles in the corpus to the training, validation, and test data.
The data sizes were 10,000 articles, 1,000 articles and 1,000 articles, respectively.
We then assigned named entity tags (LOC and PER) to each data point using the Stanford Named Entity Recognizer~\cite{Finkel:05}.
We fixed the N-gram order N to 10. 
TABLE~\ref{tab:datasets} shows the dataset descriptions\footnote{We also experimented with $\text{N=2}$ setting; however, the results were similar to $\text{N=10}$ setting. Therefore, we only show the results for $\text{N=10}$.}.
As shown in this table, the number of 10-gram types is close to the total frequency, and this implies that most of the 10-grams are infrequent.
We also confirmed that more than 99\% of the 10-gram types in the test data are zero frequencies, which are not observed in the training data.
\begin{table}[tb]
\caption{Number of Types and Total Frequency of 10-grams in Each Dataset.}
\label{tab:datasets}
\centering
	\begin{tabular}{ c  r  r  r  r  r  r } \hline
		 \multicolumn{1}{ c }{\multirow{2}{*}{Data}} & \multicolumn{2}{ c }{All} & \multicolumn{2}{ c }{Left of LOC} & \multicolumn{2}{ c }{Left of PER} \\ \cline{2-7}
		 & \multicolumn{1}{ c }{Type} & \multicolumn{1}{ c }{Freq.} & \multicolumn{1}{ c }{Type} & \multicolumn{1}{ c }{Freq.} &\multicolumn{1}{ c }{Type} & \multicolumn{1}{ c }{Freq.} \\ \hline
		Train & 3,906,050 & 3,922,930 & 62,228 & 62,532 & 66,667 & 66,766 \\
		Valid & 392,746 & 393,445 & 5,950 & 5,957 & 7,348 & 7,350 \\
		Test & 394,850 & 395,145 & 5,713 & 5,716 & 7,520 & 7,522 \\ \hline
	\end{tabular}
\end{table}

We have two experimental conditions.
The first is the named entity type, which has two choices: LOC and PER.
The second is the subset size $|\Theta_k|$ used for estimation.
In our experiments, we selected the subset $\Theta_k$ from the vocabulary $V_k$ at each position $k$ in a 10-gram.
For simplicity, we fixed the size $|\Theta_k|$ regardless of $k$.
That is, the total number of words selected from the training data is $10 \times |\Theta_k|$.
We chose one of $10^2$, $5\times10^2$, $10^3$, $5\times 10^3$, $10^4$, $5\times 10^4$ or $10^5$ as the size $|\Theta_k|$.

\subsection{Experimental Procedure}

We performed the experiment with the following procedure.
First, we decomposed all the N-grams in the training data into words and count their frequencies on the left of name entities and in the training data.
For feature selection, we selected the subset $\Theta_k$ from each vocabulary $V_k$.
Then, for each N-gram $x$ in the test data, we estimated
\begin{align}
r(x) = \frac{p(x \mid c_{\text{NE}})}{p(x \mid \bar{c}_{\text{NE}})},
\end{align}
where $c_{\text{NE}}$ is the class label assigned to $x$ that occurs on the left of named entities, and $\bar{c}_{\text{NE}}$ is the class label assigned to $x$ that occurs outside the left of the named entities in the training data.
$r(x)$ is estimated using $\Theta_k$s or $V_k$s.
We determined that the larger the estimate, the more likely that $x$ occurs on the left of the named entities.

Finally, we evaluated the performance of estimators.
We ranked N-grams in descending order of estimates and judge the top 8,000 N-grams as correct or incorrect.
We consider $x$ that occurs at least once on the left of named entities in the test data to be correct, and the others to be incorrect.
We calculated F1-measure using the judgment results.
We depict a Rank-Recall curve for the estimator with the highest F1 value.
This curve is drawn on a graph, where the horizontal axis is the rank of $x$, and the vertical axis is the recall.
The slope of the straight line connecting a certain point on the curve from the origin of the graph is proportional to the precision at point.
Precision and recall are defined as
\begin{align}
{\text{Precision}} &= \frac{|\{x \mid x \in R\}|}{|\{x\}|}, \quad {\text{Recall}} = \frac{|\{x \mid x \in R\}|}{|R|},
\end{align}
respectively.
$R$ is the N-gram set that occurs on the left of the named entities in the test data.
We measured the running time and memory usage, and compared the differences with and without feature selection.

\subsection{Comparison Estimators}\label{sec:comparison_methods}

To verify the effectiveness of feature selection, we used the following two baselines.

\textbf{All used ($\lambda = 0$):}\quad
This estimator does not use both regularization and feature selection.
This is equivalent to Eq.($\ref{eq:proposed}$) with $\lambda=0$ and $w_{k(m)}=1$.

\textbf{All used ($\lambda^*$):}\quad
This estimator does not use feature selection.
This is equivalent to Eq.($\ref{eq:proposed}$) with $w_{k(m)}=1$.
We describe how to determine $\lambda^*$, which is the optimal value of $\lambda$.

We also prepared the following score functions and compared the behaviors of our estimators using them.
From the vocabulary $V_k$, we selected $|\Theta_k|$ words with high scores to determine $w_{k(m)}$ in Eq.(\ref{eq:weight}).
The probabilities in the functions are estimated as the relative frequencies.

\textbf{Random:}\quad
We randomly selected $|\Theta_k|$ words from $V_k$.

\textbf{TF:}\quad
We selected high-frequency $|\Theta_k|$ words from $V_k$.

\textbf{CET:}\quad
The expected cross entropy for text (CET)~\cite{Mladenic:99} is defined as
\begin{align}
{\rm CET} =& \sum_{c_{lb}\in\{c, \bar{c}\}} p(t_{k(m)}, c_{lb}) \log \frac{p(t_{k(m)}, c_{lb})}{p(t_{k(m)})p(c_{lb})},
\end{align}
where $t_{k(m)}$ is the word of the $m$-th type at the $k$-th position of an N-gram.

\textbf{$\chi^2$:}\quad
The chi-square statistic~\cite{Yang:97} is defined as
\begin{align}
\chi^2 = \frac{\left[p(t_{k(m)}, c)p(\bar{t}_{k(m)}, \bar{c}) - p(t_{k(m)}, \bar{c})p(\bar{t}_{k(m)}, c)\right]^2}{p(t_{k(m)}, c)p(t_{k(m)}, \bar{c})p(\bar{t}_{k(m)}, c)p(\bar{t}_{k(m)}, \bar{c})}.
\end{align}

\textbf{GSS:}\quad
The GSS coefficient~\cite{Galavotti:00} is defined as
\begin{align}
{\rm GSS} = p(t_{k(m)}, c)p(\bar{t}_{k(m)}, \bar{c}) - p(t_{k(m)}, \bar{c})p(\bar{t}_{k(m)}, c).
\end{align}

In the comparison estimators excluding all used ($\lambda = 0$), we need to tune the regularization parameters.
For each estimator, we considered the validation data as the test data and calculated the F1 value using the top 8000 N-grams in descending order of estimates as $\lambda$ was changed to $10^{-9},\ 10^{-8},\ldots,\ 10^{-1}$.
Then, we set the value with the highest F1 value as $\lambda^*$.

\subsection{Experimental Results}

\begin{figure}[tb]
	\begin{minipage}[b]{1 \linewidth}
		\centering
		\includegraphics[keepaspectratio, scale=0.41]{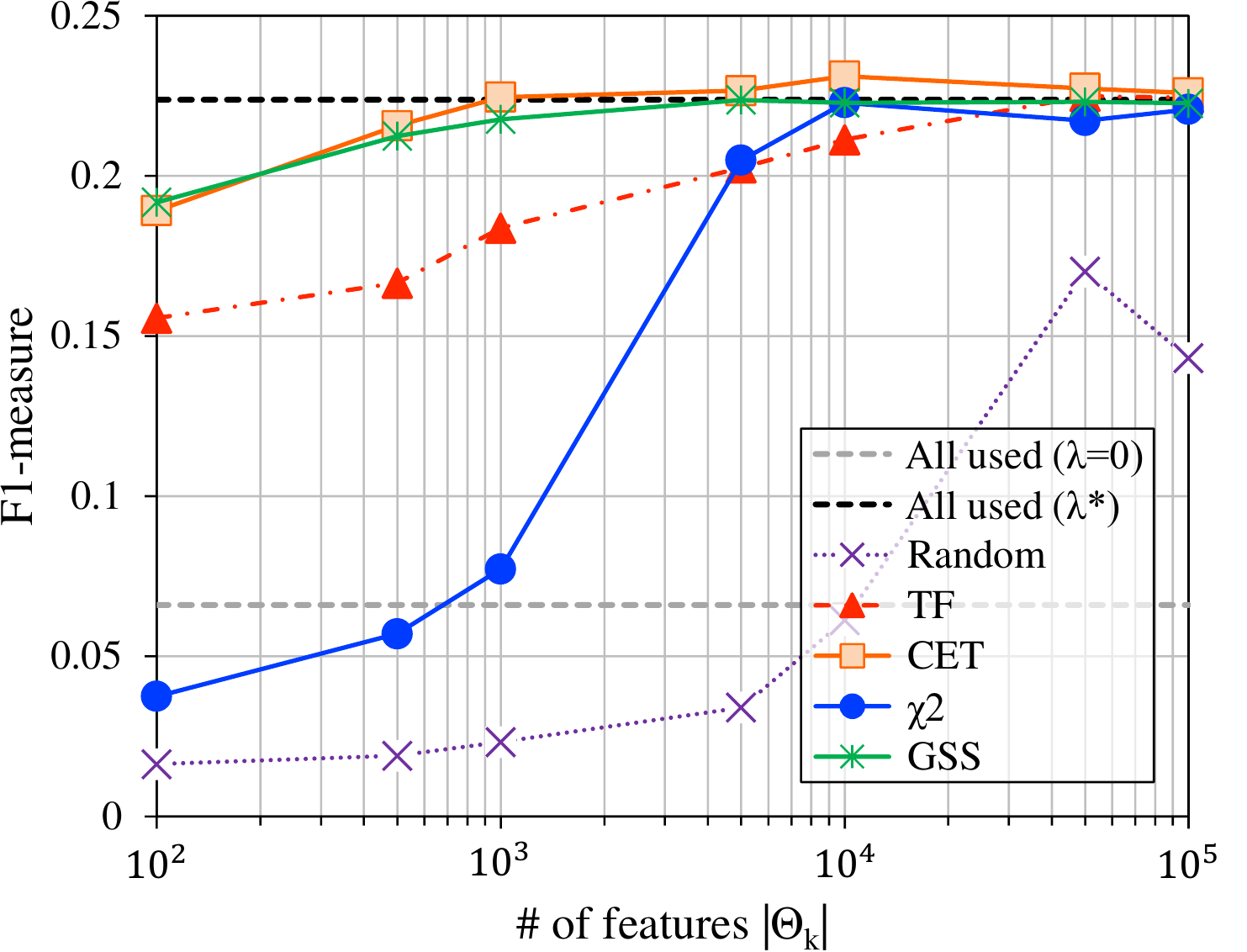}
		\subcaption{LOC}\label{fig:F1_LOC_10}
	\end{minipage} \vspace{0.0005cm} \\
	\begin{minipage}[b]{1 \linewidth}
		\centering
		\includegraphics[keepaspectratio, scale=0.41]{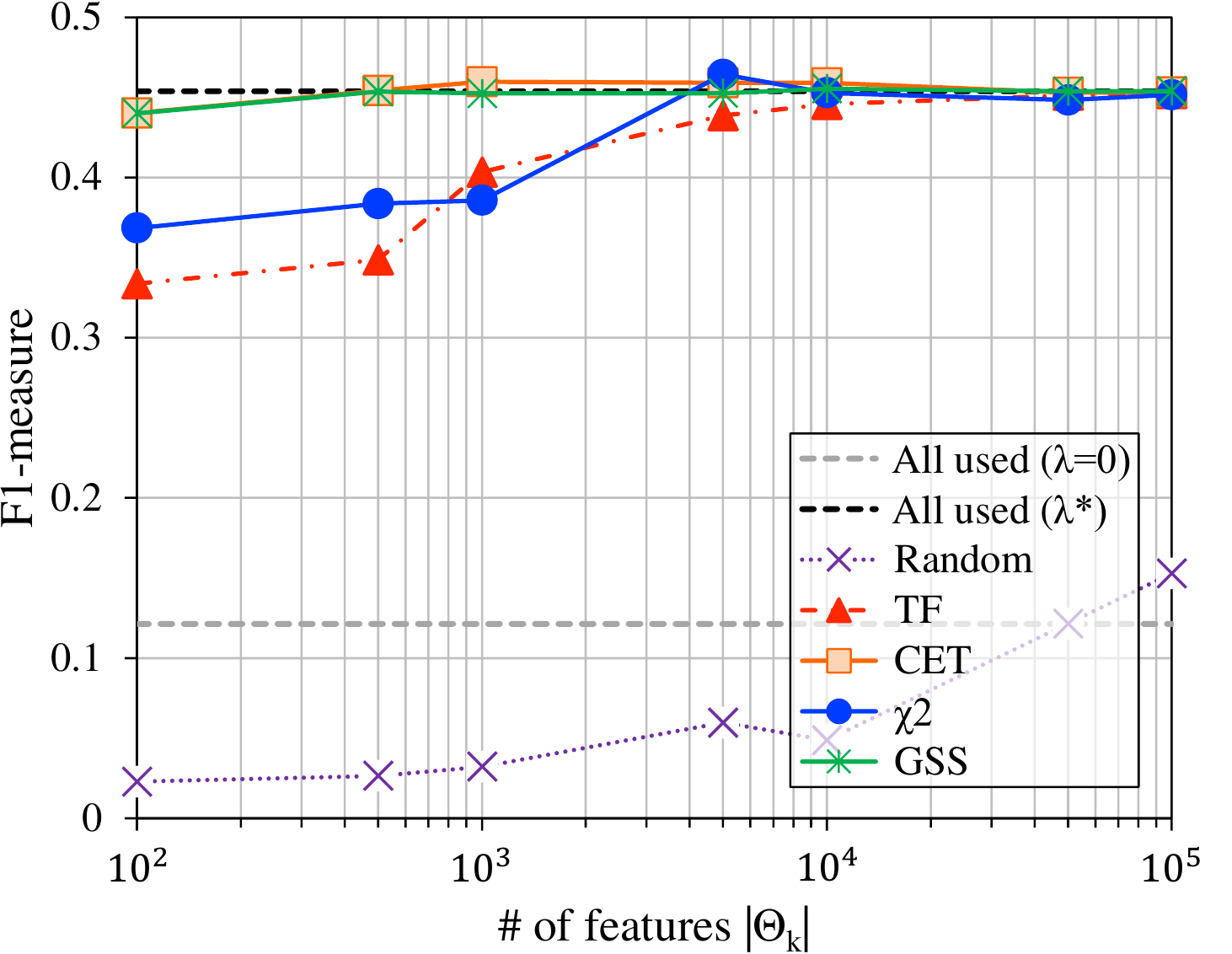}
		\subcaption{PER}\label{fig:F1_PER_10}
	\end{minipage}
	\caption{F1 values for each estimator.}\label{fig:F1}
\end{figure}
Figs.~\ref{fig:F1_LOC_10} and~\ref{fig:F1_PER_10} show the F1 values for each estimator.
The horizontal axis of each graph is the number of words $|\Theta_k|$ selected from the vocabulary $V_k$, and the vertical axis represents the F1 value for $|\Theta_k|$.
The estimator with the highest F1 value was the best in terms of the prediction accuracy.
First, focusing on the two baselines, all used ($\lambda^*$), which use the regularization parameter, has larger F1 values than all used ($\lambda = 0$), which do not.
This result suggests that the regularization parameter is effective.
Comparing Figs. \ref{fig:F1_LOC_10} and \ref{fig:F1_PER_10}, we found that the F1 values of PER contexts are almost twice as high as those of LOC contexts, indicating that PER contexts are more predictable than LOC contexts.
Next, we focus on five proposed estimators with different score functions.
Among these, Random stands out for low F1 values, and TF has low F1 values for small $|\Theta_k|$.
Therefore, it is important to select words that contribute to LR estimation.
To achieve this, we used three common score functions.
For small $|\Theta_k|$, $\chi^2$ has low F1 values; however, for $|\Theta_k|=10^4$, it has the highest F1 value in the PER case, as shown in Fig.~\ref{fig:F1_PER_10}.
While this result suggests $\chi^2$ as a potentially useful function, it also clarifies that $\chi^2$ is susceptible to adverse effects due to low frequencies.
In contrast, CET and GSS maintain stable F1 values, and for $|\Theta_k|=10^3$ or more, their F1 values are equal to or higher than those of all used ($\lambda^*$), which use all words.
Therefore, we consider them as effective score functions.

\begin{figure}[tb]
	\begin{minipage}[b]{1 \linewidth}
		\centering
		\includegraphics[keepaspectratio, scale=0.41]{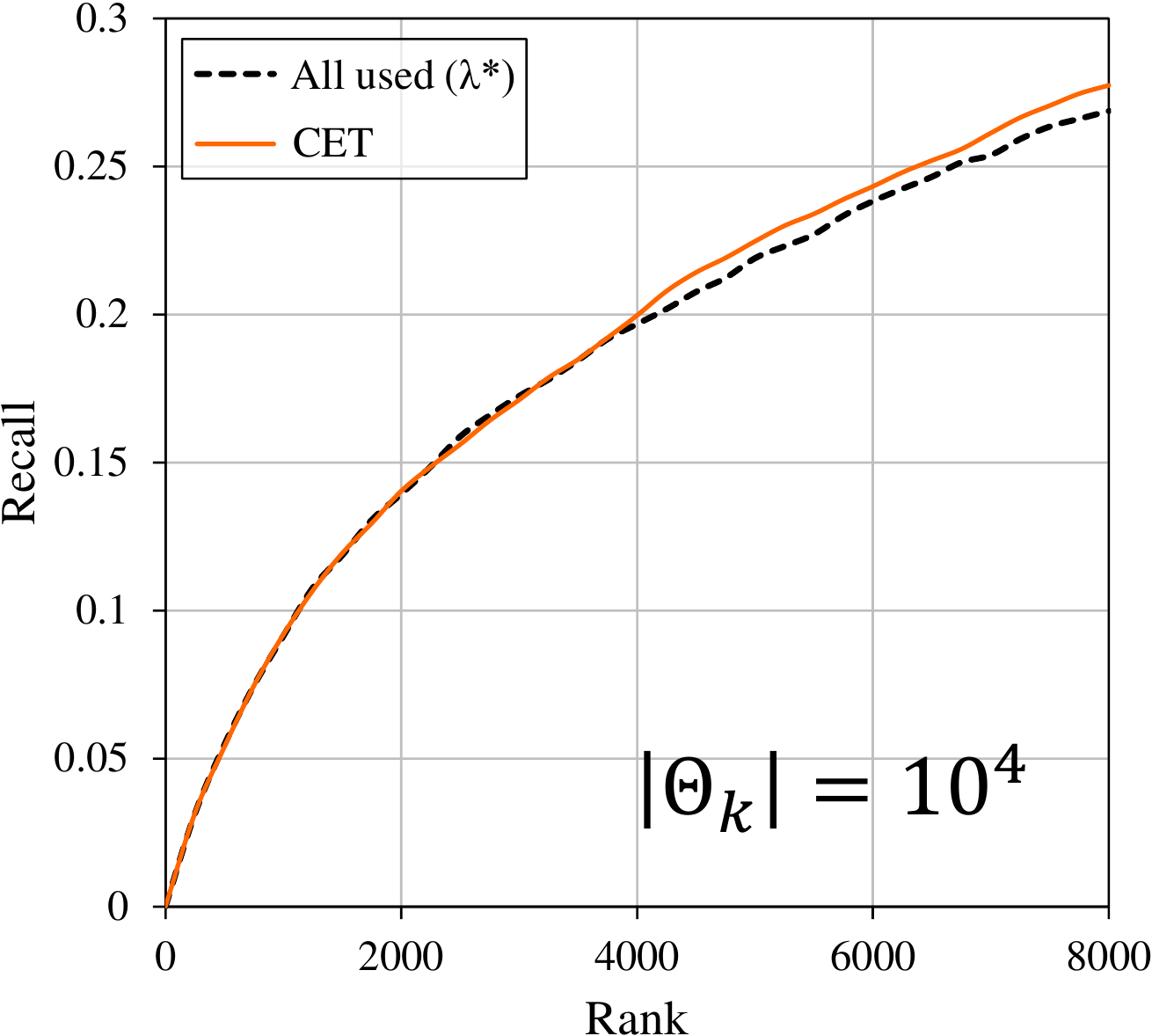}
		\subcaption{LOC}\label{fig:Recall_LOC_10}
	\end{minipage} \vspace{0.0005cm} \\
	\begin{minipage}[b]{1 \linewidth}
		\centering
		\includegraphics[keepaspectratio, scale=0.41]{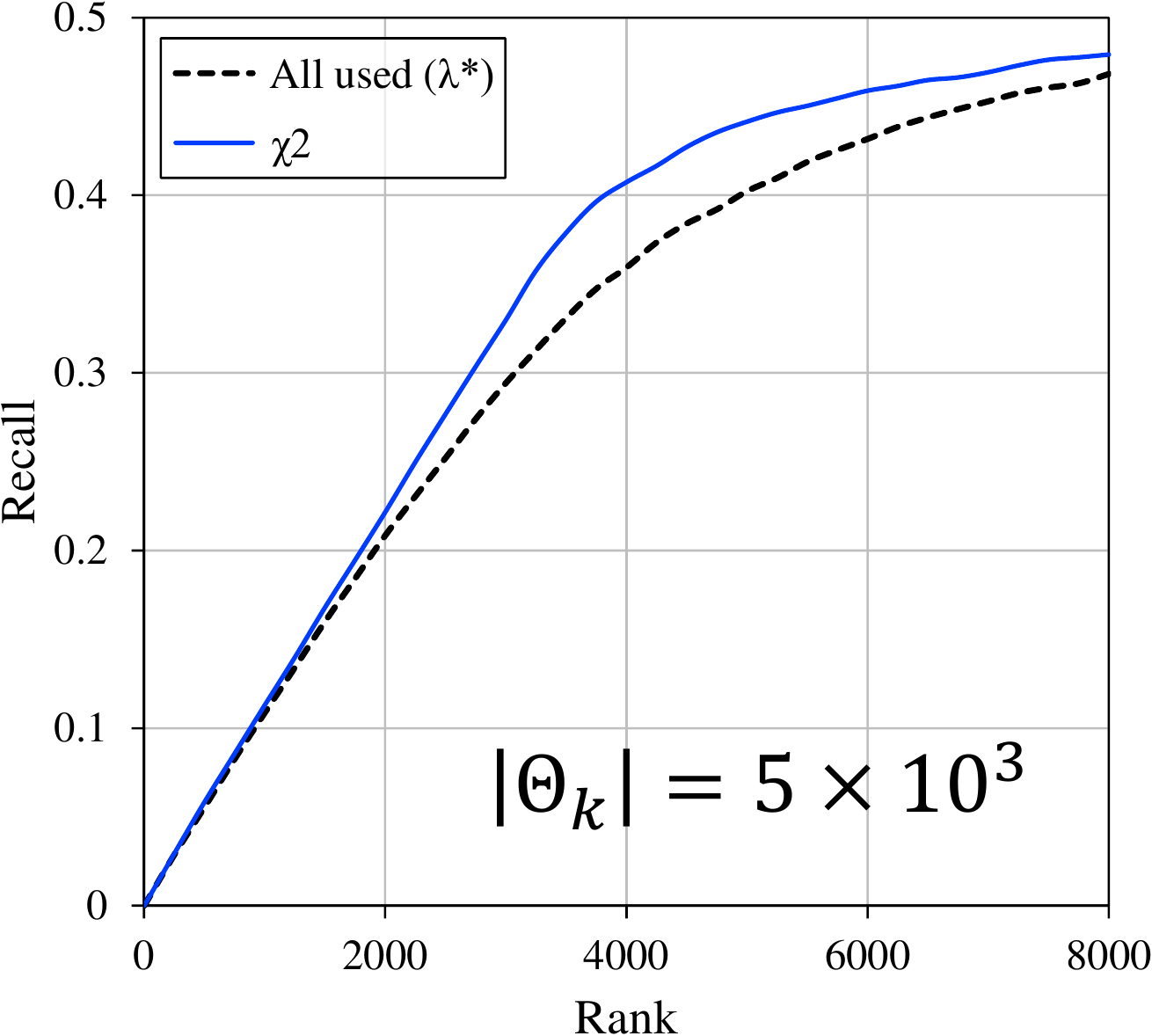}
		\subcaption{PER}\label{fig:Recall_PER_10}
	\end{minipage}
	\caption{Rank-Recall curves.}\label{fig:Recall}
\end{figure}
We compared the estimators with the highest F1 values in Figs.~\ref{fig:F1_LOC_10} and~\ref{fig:F1_PER_10} with all used ($\lambda^*$) with respect to prediction accuracy, running time, and memory usage.
For comparison, we chose $|\Theta_k|$s, which showed the highest F1 values in Figs.~\ref{fig:F1_LOC_10} and~\ref{fig:F1_PER_10}, respectively.
The running time is the time it takes to estimate the LRs for all 10-grams in the test data.
We did not include the time required to obtain a word subset in the running time because the process is performed only once and the subset can be reused for estimation.
Memory usage is the amount of memory required to store all the word frequencies from the training data.
We derived running time and memory usage by calculating the arithmetic mean of 10 runs.
Figs.~\ref{fig:Recall_LOC_10} and~\ref{fig:Recall_PER_10} show Rank-Recall curves for top 8,000 N-grams.
The Rank-Recall curve of CET is almost identical to that of all used($\lambda^*$), and the curve for $\chi^2$ has a larger slope than that for all used ($\lambda^*$) up to approximately the top 4,000.
Furthermore, $\chi^2$ maintain high recall even after top 4,000.
Thus, we confirmed that feature selection is superior in terms of prediction accuracy.
Moreover, in TABLE~\ref{tab:LOC}, the running time of CET is reduced to approximately $1/2$ of all used ($\lambda^*$), and the memory usage is reduced to approximately $1/10$ th of it.
In TABLE~\ref{tab:PER}, the running time of $\chi^2$ is reduced to approximately $1/3$ of all used ($\lambda^*$), and the memory usage is reduced to approximately $1/10$ th of it.
Overall, we confirmed the effectiveness of our estimator.

\begin{table}[tb]
\centering
\caption{Running Times and Memory Usages (LOC).}
\label{tab:LOC}
    \begin{tabular}{l c r r} \hline
        \multicolumn{1}{c}{Estimator} & Subset Size $|\Theta_k|$ & \multicolumn{1}{c}{Time [sec]} & \multicolumn{1}{c}{Memory [byte]} \\ \hline
        All used ($\lambda^*$) & $|V_k|$\ \ (All) & 30.18 & $4.45 \times 10^8$ \\
        CET & $10^4$ & 15.95 & $2.63 \times 10^7$ \\ \hline
    \end{tabular}
\end{table}
\begin{table}[tb]
\centering
\caption{Running Times and Memory Usages (PER).}
\label{tab:PER}
    \begin{tabular}{l c r r} \hline
        \multicolumn{1}{c}{Estimator} & Subset Size $|\Theta_k|$ & \multicolumn{1}{c}{Time [sec]} & \multicolumn{1}{c}{Memory [byte]} \\ \hline
        All used ($\lambda^*$) & $|V_k|$\ \ (All) & 29.94 & $4.45 \times 10^8$ \\
        $\chi^2$ & $5 \times 10^3$ & 9.00 & $2.61 \times 10^7$ \\ \hline
    \end{tabular}
\end{table}

\section{Conclusion}

This paper proposed an LR estimator for both low- and zero-frequency N-grams.
One way to handle zero-frequencies is to treat discrete values $t_k$s in an N-gram individually and take the product of their LRs.
Additionally, by applying the estimator~\cite{Kikuchi:19}, we can provide robust estimates for low-frequency $t_k$s.
However, to treat $t_k$s individually, we must assume statistical independence between $t_k$, which does not often hold in practice.
Because this method treats a large number of $t_k$s, it also deteriorates the estimation accuracy and efficiency.
To avoid these problems, we combined the aforementioned method with a feature selection method.
In our experiments, we predicted 10-grams on the left of named entities by LR estimation.
The results suggested that the feature selection method is also effective in LR estimation.
Furthermore, we compared the widely used score functions CET, $\chi^2$, and GSS.
As a result, we found that $\chi^2$ has some problems in handling low frequencies; however, CET and GSS show good and stable performance.

\section*{Acknowledgment}

This work was supported in part by JSPS KAKENHI Grant Number JP19K12266.

\bibliographystyle{IEEEtran}
\bibliography{references}

\end{document}